\documentclass[letterpaper, 10 pt, conference]{ieeeconf}  

\IEEEoverridecommandlockouts                              

\usepackage{graphics} 
\usepackage{epsfig} 
\usepackage{times} 
\usepackage{amsmath} 
\usepackage{amssymb}  
\usepackage{multirow}
\usepackage{color,soul}
\usepackage{algorithm}
\usepackage{algpseudocode}

\usepackage{hyperref}
\usepackage[comma,numbers]{natbib}

\DeclareMathOperator*{\argmin}{arg\,min}

\title{\LARGE \bf
Reachability Aware Capture Regions with Time Adjustment and Cross-Over for Step Recovery}

\author{Robert Griffin$^{1,2,3}$, James Foster$^{2}$, Stefan Fasano$^{1}$, Brandon Shrewsbury$^{1,3}$, Sylvain Bertrand$^{1}$
\thanks{This work was funded through ONR Grant N00014-19-1-2023, NASA Grant No. 80NSSC20M0197, and DAC Cooperative Agreement W911NF-21-2-0241. Email: \url{rgriffin@ihmc.org}.}
\thanks{$^{1}$Author is with the Florida Institute for Human and Machine Cognition.}%
\thanks{$^{2}$Author is with the University of West Florida.}%
\thanks{$^{3}$Author is with Boardwalk Robotics.}} 

\begin{document}

\bstctlcite{IEEEexample:BSTcontrol}

\maketitle
\thispagestyle{empty}
\pagestyle{empty}

\begin{abstract}
For humanoid robots to live up to their potential utility, they must be able to robustly recover from instabilities.
In this work, we propose a number of balance enhancements to enable the robot to both achieve specific, desired footholds in the world and adjusting the step positions and times as necessary while leveraging ankle and hip.
This includes improving the calculation of capture regions for bipedal locomotion to better consider how step constraints affect the ability to recover. 
We then explore a new strategy for performing cross-over steps to maintain stability, which greatly enhances the variety of tracking error from which the robot may recover.
Our last contribution is a strategy for time adaptation during the transfer phase for recovery. 
We then present these results on our humanoid robot, Nadia, in both simulation and hardware, showing the robot walking over rough terrain, recovering from external disturbances, and taking cross-over steps to maintain balance.
\end{abstract}

\section{Introduction}
\label{introduction}

The ability to maintain stability while walking is an inherent requirement for humanoid robots to successfully perform any task that uses locomotion.
Balancing is a skill at which humans are exceptional, with people capable of climbing sheer cliff faces, walking tightropes, running over stepping stones, and performing parkour. 
For bipedal balance specifically, several distinct strategies have been observed: shifting the Center of Pressure within the base of support (the ``ankle strategy") \cite{horak1986central}, generating angular momentum (the ``hip strategy") \cite{pijnappels2010armed}, and changing the base of support (the ``step strategy") \cite{maki1997role}. 
Utilizing these individual mechanisms for balance regulation, along with additional strategies such as varying the timing of contacts, is essential for humanoid robots to properly navigate the world in a stable fashion.

\begin{figure}
    \centering
    \includegraphics[width=0.8\columnwidth]{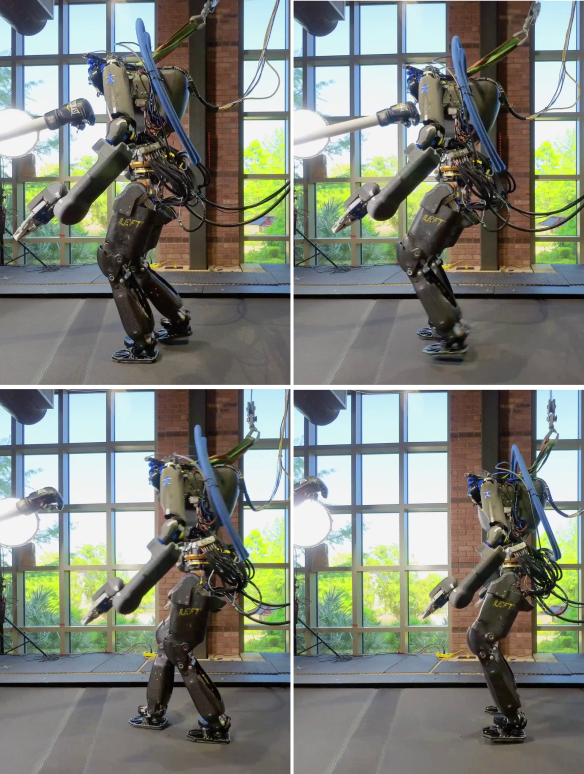}
    \vspace{-3mm}
    \caption{Our humanoid robot, Nadia, recovering from a backward shove when walking forward. 
    }
    \vspace{-8mm}
    \label{fig:nadia_push_recovery}
\end{figure}

Much of the work during the DARPA Robotics Challenge era for humanoid robots centered around properly tracking the Center of Mass (CoM) dynamics to stably execute desired footsteps \cite{Koolen_2016}.
A subsequent focus on increasing the dynamic capabilities of humanoids and closing the gap with their biological counterparts has lead to the development of a number of strategies for step adjustment for stability.
Many approaches have leveraged reduced order models to compute the best step for stability, such as the Linear Inverted Pendulum (LIP) \cite{feng2016robust}, the Instantaneous Capture Point (ICP) \cite{pratt2006capture, pratt2012capturability}, the Divergent Component of Motion (DCM) \cite{griffin2017walking, englsberger2017smooth,khadiv2020walking}, and more recently, the Angular LIP (ALIP)\cite{gong2021one}.
Model predictive control (MPC) has also become a popular technique to encode this step adjustment, leveraging reduced order models for simplicity \cite{di2018dynamic} but also using the full dynamics \cite{meduri2023biconmp}.

However, algorithms designed to leverage step adjustment often consider simple constraints on the adjustment, with little attention paid to the option for cross-over. 
One unique approach proposes uses Linear Temporal Logic to determine where to step from the LIP dynamics, including possible cross-over, and then full body kinematic optimization to avoid collisions \cite{gu2022reactive}.
Other works using MPC can generate footsteps that \textit{may} be capable of emergent cross-over due to their ability to avoid self-collisions \cite{khazoom2022humanoid}.
Alternatively, the reachable regions with cross-over have been used in a convex MPC by decomposing the non-convex region into convex regions, which are then handled as feasibility constraints in the MPC \cite{habib2022handling}.
This MPC also includes ankle torques for shifting the robot's weight.
Notably, none of these MPC-based works have demonstrated successful results on a hardware platform.
This leaves a gap in the understanding of the benefits of cross-over recovery mode is for bipedal stability. 
In fact, there's little understanding of how the inclusion of cross-over may affect stability on robots. 
Recent human studies, however, have found a strong link between the use of cross-over stepping and degradation of tracking (via disturbance) towards the end of the swing phase, indicating this recovery mode (along with jumping) is a significant mechanism employed by people\cite{leestma2023linking}.

While less explored than step adjustment strategies, step timing adaptation has garnered significant recent interest. 
However, the nonlinearity of the system dynamics with respect to time make this challenging.
It has been shown to be sufficient to vary only the next step time and position in a simple walking pattern generator with a point foot to maintain stability \cite{khadiv2020walking}, demonstrating the power of time as a control input.
The simple point foot allows treating the nonlinear term in the  dynamics as a scalar function of time. 
Timing has also been included as a decision variable in MPC \cite{aftab2012ankle, kryczka2015online, carpentier2016versatile}.
However this results in a non-convex and nonlinear problem that can be challenging to solve on real hardware with computational constraints.
The MPC problem can also be posed using a time-optimal parameterization \cite{caron2017make}, where, instead of formulating the timing between control ticks using direct transcription, the MPC parameterizes the resulting trajectory as a function of time.
Contact timing optimization can be avoided by using contact implicit trajectory optimization, where the actual event of contact is encoded in complimentarity conditions for interaction forces with the environment \cite{cleac2021fast,  posa2014direct}.
Outside of MPC, explicitly considering step timing has been limited to \textit{swing} time, with little-to-no attention provided to the effects of varying \textit{transfer} time, which is often omitted in controllers based on point foot models \cite{khadiv2020walking}.

In this work, we propose a number of enhancements that enable the robot to achieve both specific  footholds in the world while also adjusting the step positions and times as necessary. 
First, we present a novel controller for momentum shaping using the ankle and hip strategy in a compact QP that is decoupled from step adjustment. 
We next present enhancements to the calculation of capture regions for bipedal locomotion to better consider how constraints on where the robot can step may affect the ability to recover. 
We then highlight a new strategy for performing cross-over steps to maintain stability, which greatly enhances the variety of tracking error from which the robot may recover.
Our last contribution is an adaptation of our previous work on adjusting the swing duration \cite{griffin2017walking} to the transfer phase, enabling the robot to effectively ``skip" this period of double support when the stability of the system is compromised.

\section{Capture Point and Capture Region}
\begin{figure}[t]
\centering
\includegraphics[width=0.7\columnwidth]{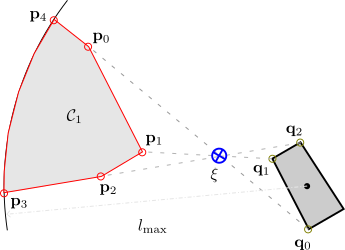}
\vspace{-4mm}
\caption{One step capture region $\mathcal{C}_1$\cite{koolen2012capturability}. 
This set can be computed using the vertices of the support foot polygon visible from the ICP, $\left(\mathbf{q}_0, \mathbf{q}_1, \mathbf{q}_2\right)$, and computing the corresponding ICP locations at $t_{\text{min}}$, $\left( \mathbf{p}_0, \mathbf{p}_1, \mathbf{p}_2 \right)$\cite{pratt2012capturability}. 
The outer bounds of $\mathcal{C}_1$ are computed as the ICP evolves over time, $\left( \mathbf{p}_3, \mathbf{p}_4 \right)$, and limited the distance to the maximum step length $l_{\text{max}}$.}
\label{fig:one_step_capture_region}
\vspace{-6mm}
\end{figure}

For locomotion onboard Nadia, we rely on controlling the instantaneous capture point (ICP) \cite{pratt2006capture}.
The ICP is a linear combination of the CoM position and velocity defined as 
\begin{equation}
    \mathbf{\xi} = \mathbf{x} + \frac{1}{\omega} \dot{\mathbf{x}},
\end{equation}
where $\mathbf{\xi}$ is the ICP position, $\mathbf{x}$ and $\mathbf{\dot{x}}$ are the CoM position and velocity, and $\omega = \sqrt{g / \Delta z_{com}}$ is the natural frequency of the inverted pendulum. 
The ICP dynamics are
\begin{equation}
    \dot{\mathbf{\xi}} = \omega \left( \mathbf{\xi} - \mathbf{r}_{\text{ecmp}}\right),
    \label{eqn:icp_dynamics}
\end{equation}
where $\mathbf{r}_{\text{ecmp}}$ is the enhanced centroidal moment pivot (eCMP) \cite{englsberger2017smooth}, which directly controls the ICP. 
The main concept of ICP control is to control the divergent dynamics with the eCMP location through either foot placement, ankle torques, or angular torque about the CoM,  so that the convergent dynamics of the CoM are indirectly stabilized. 
By placing the eCMP directly at the location of the ICP, the ICP has zero velocity, allowing the CoM position to converge over time. 

Because the ICP dynamics are first order, Eq. \ref{eqn:icp_dynamics} has the solution
\begin{equation}
    \mathbf{\xi}(t) = e^{\omega t} \left( \mathbf{\xi}_0 - \mathbf{r}_{\text{ecmp}} \right) + \mathbf{r}_{\text{ecmp}},
    \label{eqn:icp_trajectory}
\end{equation}
assuming $\mathbf{r}_{\text{ecmp}}$ is held constant throughout $t$. 
This is important, as it provides a closed form solution for where the ICP will be when the step is finished at time $T_r$. 
This, then, sets the required step location for the robot to maintain stability.

However, unless the robot has point feet and is a point mass, it is not constrained to using a fixed eCMP location, which is required by Eq. \ref{eqn:icp_trajectory}.
Instead, the robot has an allowable set of control inputs available, $\mathbf{r}_{\text{ecmp}} \in \mathcal{U}$. 
If the angular momentum rate is assumed to be zero and there is no change in height, this is equivalent to saying $\mathbf{r}_{\text{ecmp}}$ must remain within the foot, which is defined as a convex hull in practice.
In doing so, we can easily calculate the set of possible future ICP positions for all possible control inputs for time $t \in \left[ t_{min}, \infty \right)$.
This defines the \textit{one step capture region}, $\mathcal{C}_1$, shown in Fig. \ref{fig:one_step_capture_region}, as the region in which the robot must step to come to a stop in a single step.
Computing $\mathcal{C}_1$ is straightforward (see \cite{pratt2012capturability} for details), and 
can be used in some step adjustment algorithm that places the current step in $\mathcal{C}_1$.
As $C_1$ is, by construction, convex, this algorithm can be as simple as an orthogonal projection of the nominal step position onto $C_1$.

Using the ICP dynamics and a reference eCMP trajectory from from the desired steps, we can compute a reference ICP trajectory \cite{seyde2018inclusion}.
We can define a general CoM trajectory as
\begin{equation}
\begin{array}{l}
    \mathbf{x}(t) = \mathbf{c}_{0} e^{\omega t} + \mathbf{c}_{1} e^{-\omega t} + \mathbf{c}_{2} t^3 + \mathbf{c}_{3} t^2 + \mathbf{c}_{4} t + \mathbf{c}_{5},
    \end{array}
    \label{eqn:com_trajectory}
\end{equation}
which is directly the solution to the inverted pendulum dynamics \cite{kajita20013d}, assuming a  cubic eCMP trajectory.
The unknown coefficients in Eq. \ref{eqn:com_trajectory} are found by solving a constrained linear system.
These constraints can be defined as initial and final eCMP positions and velocities from the eCMP trajectory, CoM position and velocity continuity constraints at each knot point, and an initial CoM position constraint for the first segment and a terminal ICP position for the last.
This is notably similar to the approach presented in\cite{englsberger2017smooth}, but instead solving for the coefficients simultaneously as opposed to recursively, which has the benefit of flexibility constraint definition.

\section{Capture Point Control}
\label{sec:capture_point_control}
To achieve locomotion over complex terrain, one approach is to track a reference capture point location, $\mathbf{\xi}_r$, which comes from Eq. \ref{eqn:com_trajectory}. 
This can be done through momentum shaping with an ICP based on LIPM dynamics:
\begin{equation}
\dot{\mathbf{l}}_d = m \left( \omega^2 \left( \mathbf{x} - \mathbf{r}_{\text{ecmp},d}\right) + \mathbf{g} \right),
\label{eqn:momentum_law}
\end{equation}
where $\dot{\mathbf{l}}_d$ that is the net linear momentum rate objective and $\mathbf{r}_{\text{ecmp},d}$ is the desired eCMP position from the controller. 
The use of the eCMP allows the robot to use both linear  and angular momentum for balance (the ``ankle" and ``hip" strategies, respectively) \cite{englsberger2017smooth}.
The challenge is to balance the use of these tasks. 
We also specifically want to decouple this from the step adjustment mechanism, such that step adjustment does its best to maintain balance, and ICP control does its best to maintain tracking.

At the highest level, we can slightly redefine a standard proportional ICP feedback controller \cite{hopkins2014humanoid, griffin2017walking} as
\begin{equation}
\mathbf{r}_{\text{ecmp},d} = \mathbf{k}_p \left( \mathbf{\xi} - \mathbf{\xi}_r \right) + \mathbf{r}_{\text{ecmp},r}, \ \ \ \mathbf{r}_{\text{ecmp},r} = \mathbf{r}_{\text{cop},r} + \mathbf{\kappa}_r,
\end{equation}
where $\mathbf{r}_{\text{cop},r}$ and $\mathbf{r}_{\text{ecmp},r}$ are the reference CoP and eCMP positions values from  Eq. \ref{eqn:com_trajectory} and $\kappa_r$ is the reference difference between the two. 
This feedback task can be written as
\begin{equation}
        \mathbf{\delta} + \mathbf{\kappa} = \mathbf{k}_p \mathbf{\xi}_e,
        \label{eqn:feedback_task}
\end{equation}
where $\delta$ and $\kappa$ encode the CoP and eCMP feedback, respectively, and $\mathbf{\xi}_e = \mathbf{\xi} - \mathbf{\xi}_r$.
This definition results in the optimal controller output, $\mathbf{r}_{\text{cop,d}} = \mathbf{r}_{\text{cop,r}} + \mathbf{\delta}^*$ and $\mathbf{r}_{\text{ecmp,d}} = \mathbf{r}_{\text{cop,d}} + \mathbf{\kappa}^*$.


While these feedback terms can be found directly from Eq. \ref{eqn:feedback_task}, this does not provide a mechanism to balance the use of the CoP and eCMP. 
To do so, we define a basic QP as:
\begin{equation}
\begin{aligned}
    \min_{\kappa, \delta} \quad &  \left\|  \mathbf{\delta} + \mathbf{\kappa} - \mathbf{k}_p \mathbf{\xi}_e \right\|_{Q_e} + \left\|  \left( \mathbf{\delta} + \mathbf{\kappa}\right)^T \mathbf{k}_p \mathbf{\xi}_e \right\|_{Q_\perp} + \\
    & \left\| \delta \right\|_{R_\delta} + \left\| \kappa - \kappa_r \right\|_{R_\kappa} + \left\|  \mathbf{\delta} + \mathbf{\kappa} - \mathbf{\delta}_p - \mathbf{\kappa}_p \right\|_{R_p} 
    \\
\text{s.t.} \quad & \mathbf{A}_{\text{foot}} \left( \mathbf{r}_{\text{cop,r}} + \delta \right) \le \mathbf{b}_{\text{foot}}, \\
& \kappa_{\text{min}} \le \kappa \le \kappa_{\text{max}}.
\label{eqn:qp}
\end{aligned}
\end{equation}
The $Q_e$ task tries to achieve the desired feedback magnitude, the $Q_\perp$ task forces the feedback to be along the appropriate direction, $R_\delta$ and $R_\kappa$ regularize the solution, and $R_p$ penalizes deviations from the previous solution $\delta_p$ and $\kappa_p$ for consistency.
The first constraint forces the output CoP to lie within the support polygon using a half-space formulation, while the second constraint bounds the  angular momentum.

\begin{figure}[t]
\centering
\includegraphics[width=0.6\columnwidth]{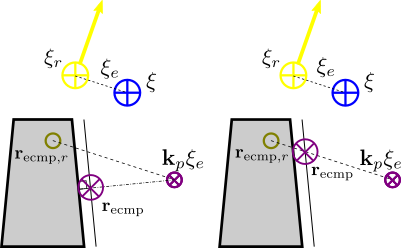}
\vspace{-2mm}
\caption{When the primary objective is achieving the desired feedback term $\mathbf{k}_p \xi_e  + \mathbf{r}_{\text{ecmp},r}$ exactly, the result orthogonally projects $\mathbf{k}_p \xi_e + \mathbf{r}_{\text{ecmp},r}$ onto the support polygon (left). 
This results in significantly different feedback direction. 
By adding the cost term with $Q_\perp$, the optimal feedback vector is parallel to $\mathbf{k}_p \xi_e$ (right).}
\vspace{-6mm}
\end{figure}

The output regulation $R_p$ is a distinct advantage of the QP-based approach. 
While feedback can be filtered using  direct rate limits or low-pass filters to the output on the feedback controller in Eq. \ref{eqn:feedback_task}, this prohibits rapid changes of force distribution for stability in the face of a sudden constraint change or disturbance.
Penalizing changes in feedback as a cost in the QP, instead, allows leveraging large feedback when necessary but avoiding small changes when not.

Once the desired eCMP is obtained, we can compute $\dot{\mathbf{l}}_d$ using Eq. \ref{eqn:momentum_law}. 
This is then supplied as a task to a whole-body controller along with other motion tasks \cite{Koolen_2016}.
In this case, a desired angular momentum rate is encoded entirely in the $\dot{\mathbf{l}}_d$ task and the other spatial acceleration tasks.
The whole-body controller balances the trade off between the necessary angular momentum rate to achieve $\dot{\mathbf{l}}_d$ and these other accelerations, allowing for perfect tracking if there are no redundant commands during nominal walking, and generating motions like pitching the torso and windmilling the arms when necessary to achieve $\dot{\mathbf{l}}_d$.

\section{Multi-Step Capture Regions}
\label{sec:multi_step_capture_regions}
\begin{figure}[t!]
    \centering
    \includegraphics[width=0.9\columnwidth]{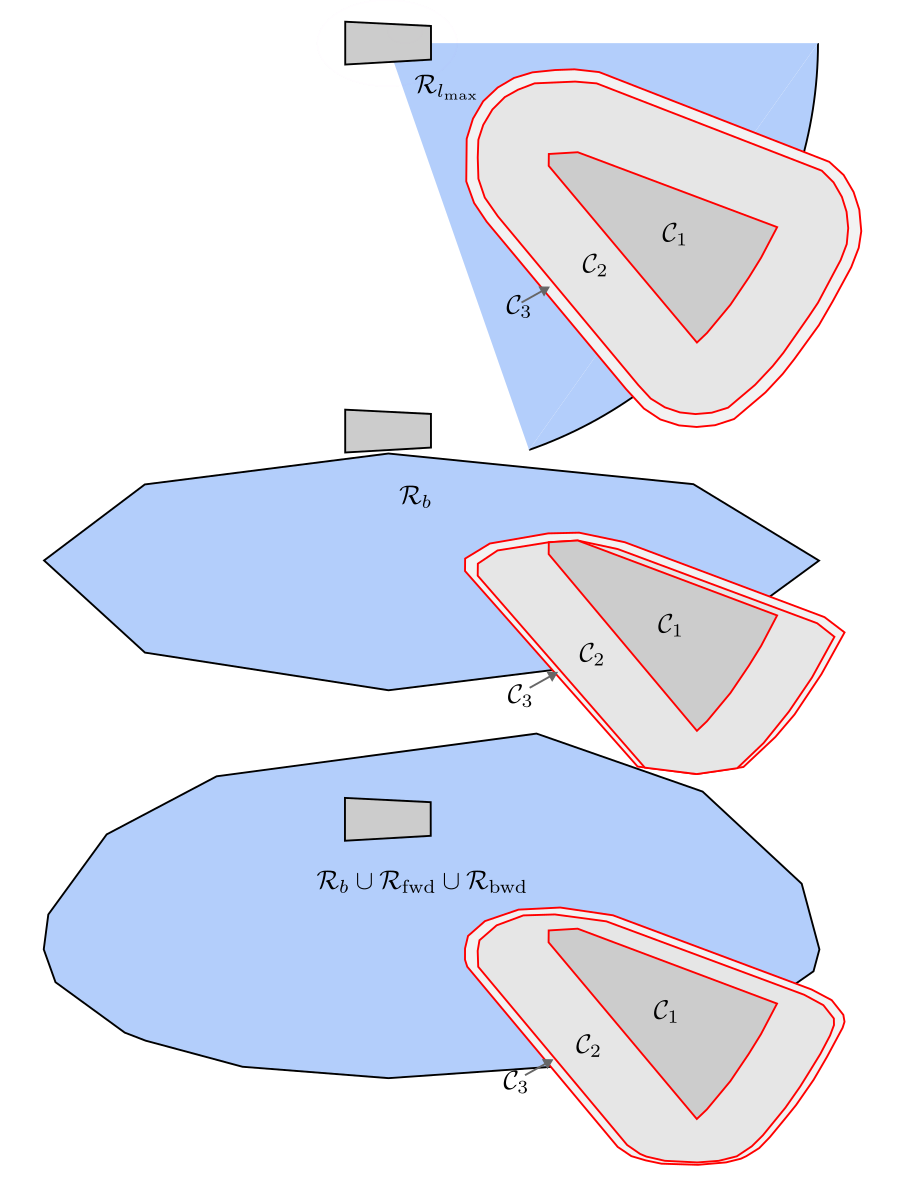}
    \vspace{-4mm}
    \caption{Multi-step capture regions for a simple, circular reachability constraint  (top), restrictive ellipsoidal reachability constraint (middle), and reachability with cross-over (bottom). As can be seen, $\mathcal{C}_2$ when not allowing crossover (middle) has almost no additional area towards the inside from $\mathcal{C}_1$. However, as cross-over is allowed, the shape of the higher order capture regions begins approaching that of circular reachability.}
    \vspace{-6mm}
    \label{fig:multi_step_capture_regions}
\end{figure}

Significantly degraded tracking from external disturbances, unstructured terrain, or slips may require multiple steps for the robot to recover.
Limiting the robot to deadbeat step adjustment using only a single step for recovery correspondingly places limits on the recoverable error.
On top of this, as the walking speed increases, the reference dynamics are often \textit{not} one step capturable, so requiring one step capturability limits the overall locomotion speed.
To address this, we consider multiple steps for recovery using multi-step capture regions \cite{koolen2012capturability}.
However, as there are reachability constraints limiting the step adjustment, these multi-step capture regions must be enhanced to consider these possibly complex restrictions.

The $N$-step capture region (referred to as $\mathcal{C}_N$) is the region in which the robot must step to avoid falling in $N$ or fewer steps. 
This region is constructed using a series of nested regions, with the innermost being $\mathcal{C}_1$, and growing increasingly larger until $\mathcal{C}_N$ is reached, with $\mathcal{C}_{N-1} \subset \mathcal{C}_N$.
By definition, after a step is taken in region $\mathcal{C}_N$, the system is then in a $N-1$ capturable state, such that the next step can be taken in region $\mathcal{C}_{N-1}$ to recover \cite{koolen2012capturability}.

To compute $\mathcal{C}_N$, we can first define a reachability constraint that restricts the position of step $n$, $\mathbf{r}_{\text{foot}, n}$, to some reachable area, $\mathbf{r}_{\text{foot}, n} \in \mathcal{R}_n$, which can be scaled in size by $\alpha$ and translated by $\delta$, $\alpha \mathcal{R} + \delta$.
From the ICP dynamics, we know that the ICP state at the end of step $n$ is 
\begin{equation}
    \mathbf{\xi}_n = e^{\omega T_s} \left( \mathbf{\xi}_{n-1} - \mathbf{r}_{\text{foot},n-1} \right) +  \mathbf{r}_{\text{foot},n-1},
\end{equation}
where $T_s$ is the step duration. As $\mathbf{r}_{\text{foot}, n} \in \mathcal{R}_n$, we can define the remaining error after step $n$ is taken, $\mathbf{\xi}_{e,n} = \mathbf{\xi}_n - \mathbf{r}_{\text{foot},n}$, as the shortest vector from $\mathbf{\xi}_n$ to $\mathcal{R}_n$, or
\begin{equation}
\mathbf{\xi}_{e,n} = \mathbf{d} \left(\mathbf{\xi}_n, \mathcal{R}_n \right).
\end{equation}
From Eq. \ref{eqn:icp_dynamics}, if the state can be captured on the $N^{th}$ step,  $\mathbf{\xi}_{e,N} = 0$, or
\begin{equation}
    \mathbf{d} \left( \xi_{e, N},  \mathcal{R}_{N}\right) = 0.
    \label{eqn:captured_definition1}
\end{equation}
As $\mathcal{R}_N$ is always defined centered at $\mathbf{r}_{foot,N}$, we can define everything assuming $\mathbf{r}_{foot,N} = 0$. From this, it follows that $\mathbf{\xi}_{e,N} = e^{\omega T_s}$. 
This transforms Eq. \ref{eqn:captured_definition1} to
\begin{equation}
    \mathbf{d} \left( \xi_{e, N-1}, e^{-\omega T_s} \mathcal{R}_{N}\right) = 0.
\end{equation}
This is equivalent to saying that, to be capturable in $N$ steps,  
\begin{equation}
\xi_{e,N-1} \in e^{-\omega T_s} \mathcal{R}_N.
\label{eqn:error_capture}
\end{equation}
This allows us to calculate the maximum amount of additional error that can be rejected by adding step $n$ for recovery, recursing back in time to the end of the current step with $e^{\omega T_s (n-1)}$.

From here, we can define $\mathcal{C}_N$ as the set of all footstep locations that drive the state, $\xi$, to $\mathcal{C}_{N-1}$.
Region $\mathcal{C}_N$, then, can be found as the set of all points $\mathbf{r}$ such that the scaled reachable region intersects the previous capture region, or
\begin{equation} 
\mathcal{C}_N = \left\{ \mathbf{r} \ | \ \left( e^{-\omega T_s \left( N - 1 \right)} \mathcal{R}_N + \mathbf{r} \right) \  \cap \  \mathcal{C}_{N-1} \right\},
\label{eqn:capture_region}
\end{equation}
where $N-1$ captures the recursive nature of this operation.
Then, if the current step is taken in the $C_N$ region, the robot can recover in $N$ steps, with each step being reachable, $\mathbf{r}_n \in \mathcal{R}_n, \forall n = 1, \dots, N$. 

This is a subtly different definition than that presented previously \cite{koolen2012capturability}, as it is designed for arbitrary reachability constraints $\mathcal{R}$.
However, if a simple reachable set is used, defined only by a maximum step length, $l_\text{max}$, where $\mathcal{R}_{l_\text{max}} = \left\{ \mathbf{r} \ | \ \| \mathbf{r} \| \le l_{\text{max}} \right\}$, the resulting capture regions are the same. 
In this case, $\mathcal{C}_N$ is simply a direct expansion of $\mathcal{C}_{N-1}$ by a distance of $e^{-\omega T_s (N-1)} l_{\text{max}}$, resulting in $\mathcal{C}_{1,2,3}$ shown on the top in Fig. \ref{fig:multi_step_capture_regions}.

\begin{figure}[b!]
    \centering
    \vspace{-6mm}
    \includegraphics[width=0.7\columnwidth]{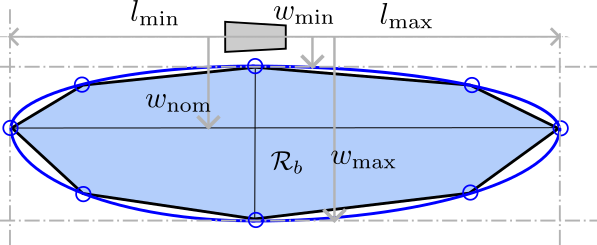}
    \vspace{-4mm}
    \caption{Reachability $\mathcal{R}_b$ can be described as an ellipse (blue line) centered about a nominal position $w_{\text{nom}}$ with width bounds $\left( w_\text{min}, w_\text{max} \right)$ and length bounds $\left(  l_\text{max}, l_\text{min} \right)$, and is approximated by an interior polygon (black line).}
    \label{fig:reachability_constraint}
\end{figure}

It should be noted that, when computing $\mathcal{C}_N$, there are additional control inputs available to the robot besides step position, namely ankle and hip torques, as well as step timing adjustment (as in \cite{griffin2017walking}) that have been ignored in our approach for calculating $\mathcal{C}_N$.
This makes the presented method an interior estimation of the true $\mathcal{C}_N$, and leads to more step adjustment on each step than may be strictly necessary for recovery.
In practice, we prefer this conservative approach, as it provides a better factor of safety for the adjustment. 
Additionally, by using more step adjustment, the robot's gait is less likely to lead to foot tipping or severe torso angles from saturating the ankle and hip control strategies, respectively.

A major limitation of the use of the simple reachability constraint $\mathcal{R}_{l_\text{max}}$, is the assumption that the robot can step anywhere within $l_{\text{max}}$ of its stance foot.
In practice, it is often useful to define some convex reachability constraint for the robot to prevent a number of undesirable events from occurring, including stepping too wide, stepping on its own feet, or colliding with the stance leg. 
To address this, we use an ellipsoidal constraint on the swing foot relative to the stance foot, defined as $\mathcal{R}_b$ and shown in Fig. \ref{fig:reachability_constraint}, which allows constraining the minimum and maximum step width and length.
We approximate this ellipse as a polygon.

\begin{figure*}[ht!]
    \centering
    \includegraphics[width=0.9\textwidth]{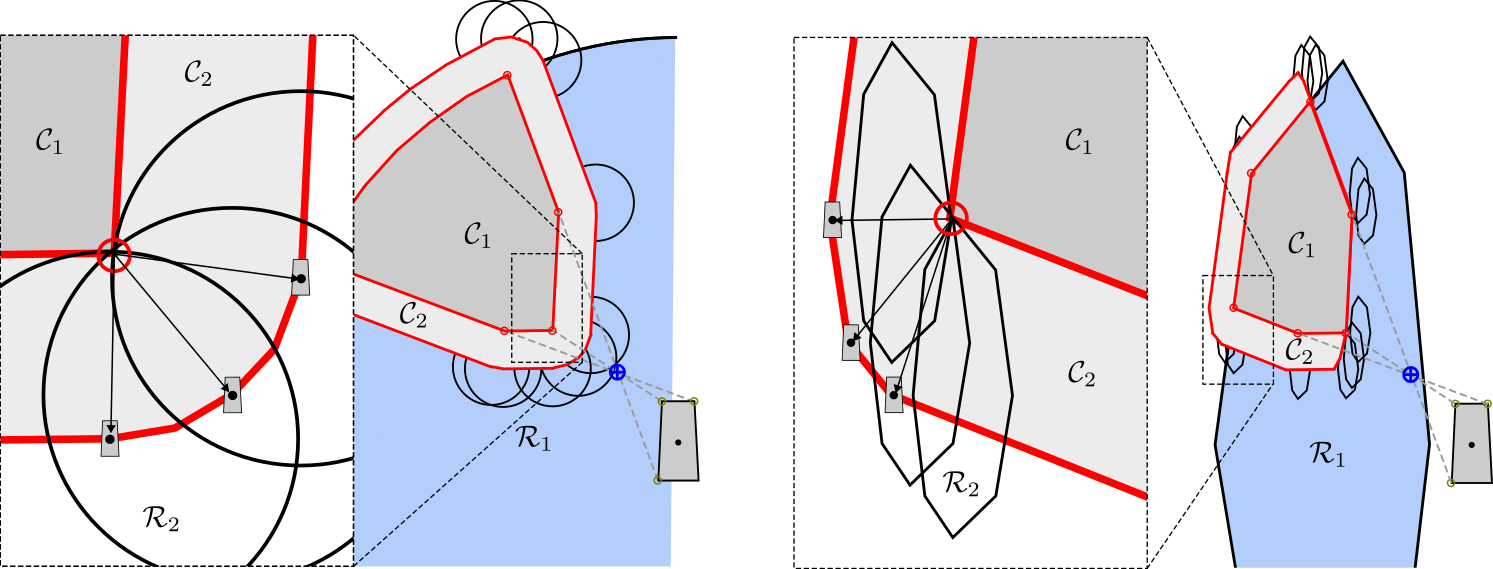}
    \vspace{-3mm}
    \caption{The expansion of $\mathcal{C}_{N-1}$ to compute $\mathcal{C}_N$ can be calculated by sweeping a time-scaled version of the reachability region, $e^{\omega (N-1) T_s} \mathcal{R}_N$, about $\mathcal{C}_{N-1}.$ 
    This is shown for both $\mathcal{R}_{l_\text{max}}$ (left), and $\mathcal{R}_b$ (right).}
    \vspace{-6mm}
    \label{fig:two_step_capture_regions}
\end{figure*}

We can then compute the capture region using this new ellipsoidal reachable region in Eq. \ref{eqn:capture_region}.
From this, the boundary of $\mathcal{C}_N$ is defined as the foot position at the intersection of the transformed $\mathcal{R}_N$ and the boundary of $\mathcal{C}_{N-1}$. 
As both $\mathcal{R}_N$ and $\mathcal{C}_{N-1}$ are polygons, $\mathcal{C}_N$ is by construction a polygon, and the outer vertices of $\mathcal{C}_N$ come from the intersection between vertices of the transformed $R_N$ and $\mathcal{C}_{N-1}$.
Algorithmically, this is equivalent to calculating $\mathcal{C}_N$ as the valid foot positions when sweeping the scaled $\mathcal{R}_N$ around the perimeter of $\mathcal{C}_{N-1}$, as shown in Fig. \ref{fig:two_step_capture_regions}.

From Fig. \ref{fig:multi_step_capture_regions}, including knowledge of this reachable set strongly affects the shape of $\mathcal{C}_N$, with the results of using $\mathcal{R}_b$ to compute the $\mathcal{C}_N$ shown in the middle.
It is apparent that, because of the minimum width constraint, additional steps do not provide much additional stability when the error is towards the inside of the foot, as recovery requires almost completely stepping in $\mathcal{C}_1$.
This can be seen by the lack of additional area in the inward direction when comparing $\mathcal{C}_2$ to $\mathcal{C}_1$ in the middle of Fig. \ref{fig:multi_step_capture_regions}.
This makes sense; the original, simple reachable set assumes the robot can take a subsequent step of width $l_\text{max}$, when in actuality, by prohibiting cross-over, the closest width for the next step is $-w_{\text{min}}$. 
The second step, then, provides no corrective control in the inward direction as the control authority (defined by $\mathcal{R}_b$) is actually \textit{negative}.
Thus, if $\mathcal{R}_{l_\text{max}}$ was used to compute $\hat{\mathcal{C}}_N$, the robot may have stepped outside the actual $\mathcal{C}_N$, leading to a fall! 

\section{Cross-Over Reachability}
\begin{figure}[b!]
    \centering
    \vspace{-6mm}
    \includegraphics[width=0.8\columnwidth]{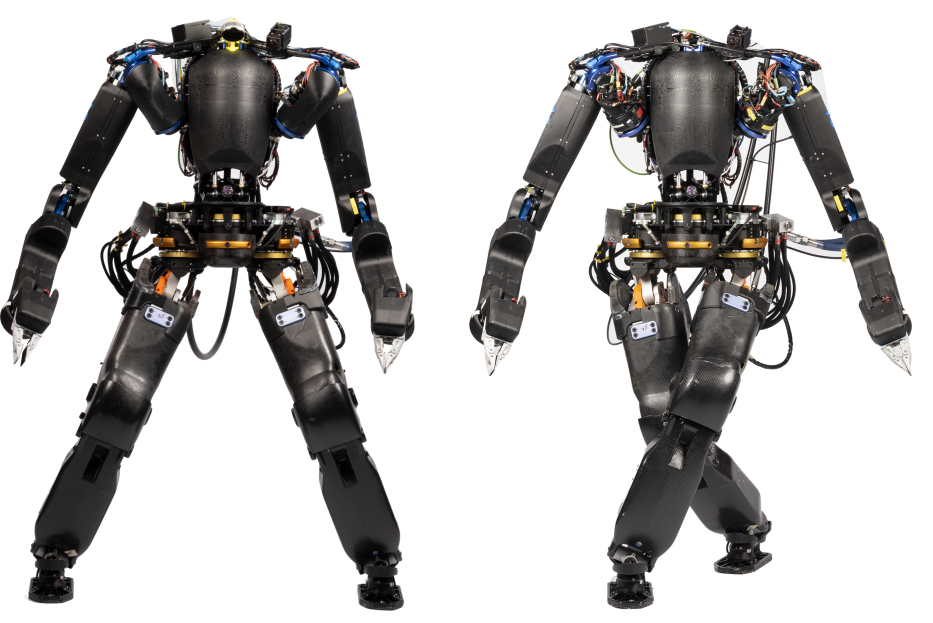}
    \vspace{-4mm}
    \caption{The hip roll range of motion on our robot, Nadia, was designed to allow the robot to perform both wide steps and cross-over steps.}
    \label{fig:nadia-crossover}
\end{figure}

While the reachability-aware capture regions provide a better model for step recovery, they do highlight the fundamental limitations encountered by preventing cross-over steps.
Our robot, Nadia, was specifically designed with a range of motion at the hip roll joint that would allow for cross-over, as shown in Fig. \ref{fig:nadia-crossover}.
To include this in our capture region calculation, however, we need to redefine the reachability constraint shown in Fig. \ref{fig:reachability_constraint}.

\begin{figure}[b!]
    \centering
    \vspace{-8mm}
    \includegraphics[width=0.9\columnwidth]{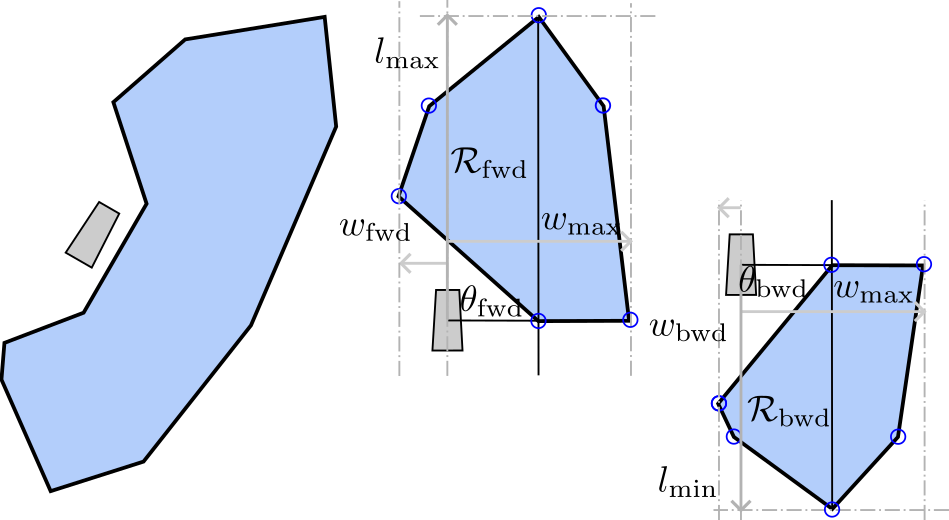}
    \vspace{-4mm}
    \caption{Reachability constraint that allows cross-over (left). We can decompose this non-convex constraint into three separate convex regions, one with forward cross-over (middle), one with backward cross-over (right), and the base region in Fig. \ref{fig:reachability_constraint}}
    \label{fig:crossover_reachability}
\end{figure}

To address this, we can create a new reachability constraint that allows for cross-over, shown in Fig. \ref{fig:crossover_reachability}.
We can define the shape of this region by imposing a maximum forward and backward cross-over distance, $w_\text{fwd}$ and $w_\text{bwd}$, as well as a cross-over angle $\theta_\text{fwd}$ and $\theta_\text{bwd}$.
This allows different cross-over amounts in the forward and backward directions, with the angle enabling stance leg collision avoidance.
However, this new reachability region is non-convex, making much of the computation and constraint formulation significantly more complex.
Instead, similar to \cite{habib2022handling}, we can decompose this non-convex region into three convex regions: one for forward cross-over $\mathcal{R}_\text{fwd}$, one for backward cross-over $\mathcal{R}_\text{bwd}$, and the original reachability constraint with no cross-over $\mathcal{R}_b$.

While this methodology is not dissimilar to previous works \cite{habib2022handling}, instead of selecting the region based on the effect when applied to a MPC, we define a set of rules to determine the appropriate constraint based on $\mathcal{C}_N$. 
The first rule is if there is overlap between $\mathcal{C}_N$ and  $\mathcal{R}_{\text{b}}$, apply this constraint for $\mathcal{R}_1$.
The second rule is, if there is no intersection between $\mathcal{R}_\text{b}$ and $\mathcal{C}_N$,  pick the region between $\mathcal{R}_{\text{fwd}}$ and $\mathcal{R}_{\text{bwd}}$ that has the most intersection with $\mathcal{C}_N$.
This can be thought of as selecting the reachability constraint that provides the largest factor of safety.
This prioritization of the base reachability constraint before using either of the cross-over constraints ensures that cross-over is only used when $\mathcal{R}_\text{b}$ is insufficient to stabilize the system.

The third and final rule is, if there is no intersection between any reachability constraint and $\mathcal{C}_N$, select the reachability constraint region that is closest to $\mathcal{C}_N$. 
From the definition of the capture region, there being no intersection between $\mathcal{C}_N$ and $\mathcal{R}_1$  strictly means that there is no way for the robot to recover, as it is impossible to step in $\mathcal{C}_N$.
However, as mentioned in Sec. \ref{sec:multi_step_capture_regions}, when we calculate $\mathcal{C}_N$, we do not consider the ``ankle" or ``hip" strategies as part of the feedback, or the ability to step more quickly. 
This means that $\mathcal{C}_N$ is a conservative estimate of the \textit{real} multi-step capture region.
Because of this, we enable the robot to continue to try to regain balance, even if there are no explicitly feasible recovery steps available.

\section{Transfer Time Adjustment}
\begin{figure}
    \centering
    \includegraphics[width=0.9\columnwidth]{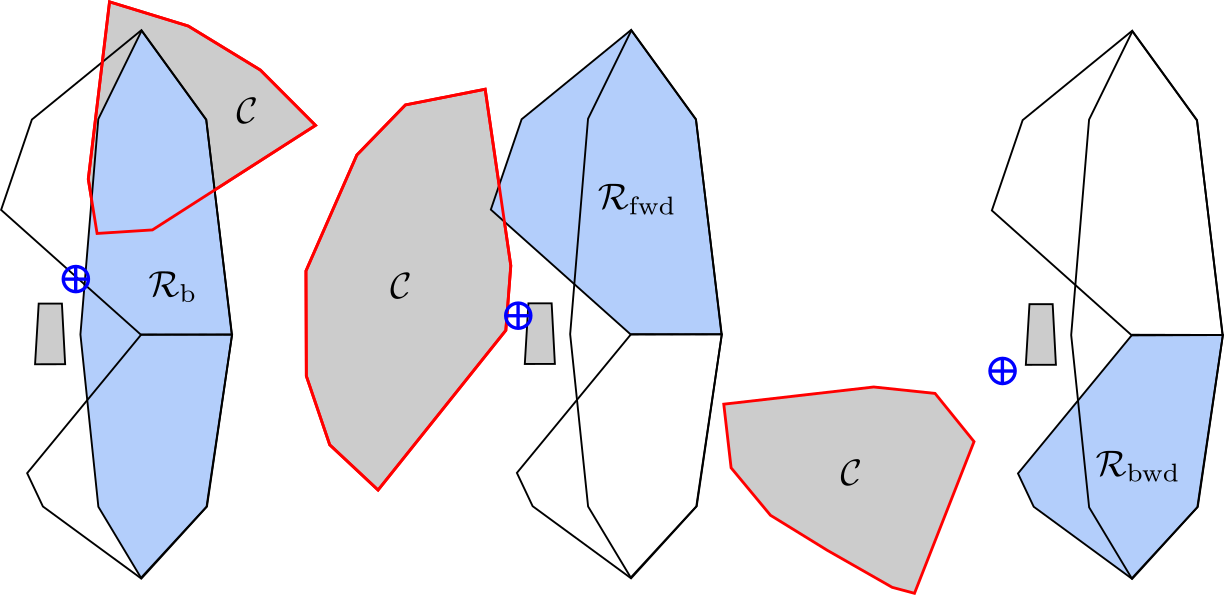}
    \vspace{-4mm}
    \caption{Illustration of the three rules for selecting the appropriate reachability constraint (blue). The first rule (left) says to select $\mathcal{R}_b$ if it has any intersection with $\mathcal{C}$. If not, select between the regions $\mathcal{R}_\text{fwd}$ and $\mathcal{R}_\text{bwd}$ the region that has the most intersection with $\mathcal{C}$ (middle). If there is no intersection with any of these regions, select the closest to $\mathcal{C}$ (right).}
    \vspace{-6mm}
    \label{fig:enter-label}
\end{figure}

An additional control mechanism is to modify the swing duration to allow the robot to take the current step more quickly \cite{griffin2017walking,khadiv2020walking}. However, when a disturbance requires more than a single step to recover, relying only on speeding up the swing is no longer sufficient when the gait includes a double-support ``transfer" phase.
Many robots that rely primarily on step-adjustment for stability avoid the complexity of the added transfer phase by omitting it from the gait entirely\cite{khadiv2020walking, gong2021one}.
For navigating rough terrain, though, the inclusion of a transfer phase can be important to facilitate proper weight shifting to unload the feet.
When performing step adjustment, however, the goal is to change the base of support as quickly as possible to get the eCMP into the necessary position for the robot to become capturable. 
Any included transfer phase delays this shifting, making the robot less stable.
Thus, it is necessary to determine an appropriate strategy for ``speeding up" the transfer phase.

Mathematically, time adjustment can be described as finding the time $t^*$ that minimizes the difference between the reference capture point $\mathbf{\xi}_r(t^*)$ from Eq. \ref{eqn:com_trajectory} 
 and the current state $\mathbf{\xi}$.
This is equivalent to solving
\begin{equation}
    t^* = \argmin_\tau \left\| \mathbf{\xi}_r(\tau) - \mathbf{\xi} \right\|,
\label{eqn:time_optimization}
\end{equation}
but noting that $\mathbf{\xi}_r(\tau)$ is generally highly nonlinear.

If the eCMP is assumed to be a constant value, the dynamics of Eq. \ref{eqn:com_trajectory} simplify to those in Eq. \ref{eqn:icp_trajectory}, and methods such as the one previously used in \cite{griffin2017walking} become possible. 
In this approach, we know that the ICP evolves along a line from its current value at $\xi_r = \xi(t)$ to the final value at $\xi_T = \xi(T_s)$. 
We then know that the closest ICP along the plan, satisfying Eq. \ref{eqn:time_optimization}, will occur at the orthogonal projection of $\mathbf{\xi}$ onto the line $\mathbf{\xi}_T - \mathbf{\xi}_r$, resulting in $\mathbf{\xi}_p$. 
This then leads to the time adjustment
\begin{equation}
    \Delta t = \frac{1}{\omega} \log \left( \frac{\xi_p - \mathbf{r}_{\text{ecmp},r}}{\xi_r - \mathbf{r}_{\text{ecmp},r}}\right),
    \label{eqn:time_delta}
\end{equation}
where $t^* = t + \Delta t$.

This assumption of constant eCMP value is not representative of the transfer phase, however, where the robot is shifting the weight from one foot to the next. 
While transfer still has a closed form definition from Eq. \ref{eqn:com_trajectory} (see  \cite{englsberger2017smooth}), the optimization for time becomes much more challenging.
Instead, in this work we choose to simply apply a discount rate $\gamma$ to the adaption in Eq. \ref{eqn:time_delta}, making the update law $t^* = t + \gamma \Delta t$.
By setting $\gamma$ to be sufficiently small, this result converges to the actual $t^*$ as the adjustment is solved iteratively from one control tick to the next. 

Even when not performing step recovery, the ability to adjust the time in transfer has been found to be beneficial. 
When using just the feedback controller in Sec. \ref{sec:capture_point_control}, if the actual ICP is leading the reference ICP, the robot will heavily shift the eCMP towards to the upcoming foot to ``brake" the dynamics and converge back to the plan. It will then shift the eCMP back to the trailing foot and ``push" to resume the plan.
If, instead, the time is simply adjusted forward, the amount of feedback is decreased and this ``brake then push" phenomena is avoided. 
As we are also applying the time adjustment law in \cite{griffin2017walking} to the swing phase, this type of control becomes akin to using the actual ICP position as a monotonically increasing phase variable for determining $\mathbf{\xi}_r$.

\section{Results}
\begin{figure}[t]
    \centering
    \includegraphics[width=0.95\columnwidth]{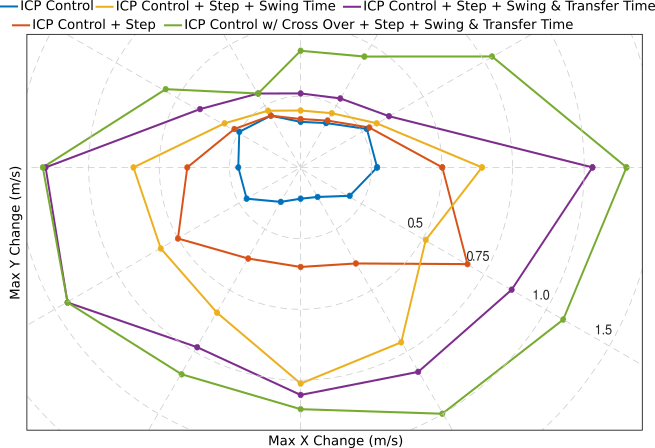}
    \vspace{-3mm}
    \caption{Maximum simulated disturbance bounds, in change in CoM velocity, that the robot can recover from.}
    \vspace{-6mm}
    \label{fig:sim_recoverable_disturbances}
\end{figure}

We used the presented walking controller in both simulation and hardware experiments on our robot, Nadia.
For our first experiment, to investigate how the different stabilization mechanisms affect the robot's balance, we ran simulations with different feedback mechanisms enabled in the following combinations:
only ICP control from Sec. \ref{sec:capture_point_control}; ICP control and step adjustment; ICP control with step and swing duration adjustment; ICP control with step, swing, and transfer duration adjustment; and all mechanisms--ICP control, step with cross-over, swing, and transfer duration adjustment.
The robot was told to walk in place, and then a disturbance was applied to the pelvis at 0.25\% of the way through swing with the right foot for 0.1s.
We then recorded the magnitude of the maximum recoverable disturbance, and mapped this to a change in velocity to normalize for push duration and robot weight. The results are shown in Fig. \ref{fig:sim_recoverable_disturbances}.
The step timings were 0.7 s in swing and 0.3 s in transfer, similar to what is used on hardware. For reachability, $l_\text{max} = l_\text{min} = 1.0m, w_\text{min} = 0.125m, w_\text{max} = 0.8m, w_\text{nom} = 0.25m, w_\text{fwd}=0.1m, w_\text{bwd} = -0.05m, \theta_\text{fwd} = 20^\circ$ and $\theta_\text{bwd}=30^\circ$.
As can be seen in Fig. \ref{fig:sim_recoverable_disturbances}, the addition of each new control mode all increase the disturbance that the robot can recover from. 
The only exception is that including transfer adaptation does not seem to increase recovery when the push is slightly towards the outside and mostly forward.
When comparing pushing towards the inside to outside, only ICP control alone shows greater recovery, likely due to the increased support polygon from the next step.
From these plots, it is easy to see the expansion of the recovery region via cross-over steps results in a greatly increased ability to recover from inward disturbances.
The exception is when they're slightly backward, which would require stepping directly through the stance foot, which is not allowed in the reachability constraint.

\begin{figure}[t]
    \centering
    \includegraphics[width=0.9\columnwidth]{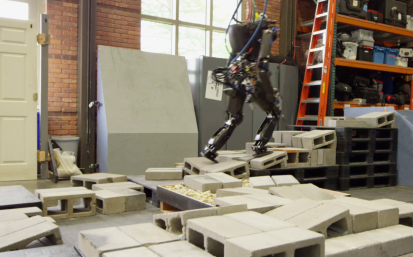}
    \vspace{-2mm}
    \caption{Nadia walking to the top of a cinder block pile using the balance algorithm outlined in Sec. \ref{sec:capture_point_control} with no step adjustment.}
    \vspace{-6mm}
    \label{fig:nadia_rough_terrain}
\end{figure}

As a feasibility proof for the ICP controller in Sec. \ref{sec:capture_point_control} to stabilize the robot on rough terrain, we set up a cinder block field, shown in Fig. \ref{fig:nadia_rough_terrain}, for the robot to traverse. 
Footsteps were provided by an operator for the robot to execute. 
As shown in Fig. \ref{fig:nadia_rough_terrain} and in the supplemental video, the robot was able to stably execute these steps to climb this terrain.

\begin{figure}[t]
    \centering
    \includegraphics[width=0.65\columnwidth]{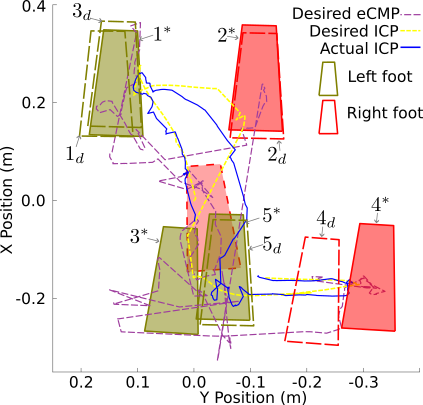}
    \vspace{-2mm}
    \caption{Data from the push shown in Fig. \ref{fig:nadia_push_recovery}. Desired steps are dashed, achieved steps are solid and filled and shown as $[\cdot]^*$. The initial right stance foot has a finer dash.}
    \vspace{-6mm}
    \label{fig:nadia_data_treadmill_stepadjustment}
\end{figure}

In the first push recovery hardware experiment, we commanded the robot to walk forward, and applied disturbances to the torso, as shown in Fig. \ref{fig:nadia_push_recovery}.
In this experiment, the robot is using a swing duration of 0.6s and transfer duration of 0.2s, and the disturbance shown is applied at the very end of the right swing. 
The subsequent transfer duration is executed in 0.03s, the next swing in 0.4s, the following transfer in 0.17s. 
At this point, the robot starts to stabilize, executing the second step in 0.49s.
The robot made a backward step adjustment of approximately 30cm to balance.
The step adjustment employed by the robot is shown in Fig. \ref{fig:nadia_data_treadmill_stepadjustment}.
This also shows how the robot moved the eCMP in the base of support to try to recover balance.
This demonstrates the combined techniques of swing duration, transfer duration, and step position adjustment are effective for regaining balance.

\begin{figure}[b]
    \centering
    \vspace{-6mm}
    \includegraphics[width=0.75\columnwidth]{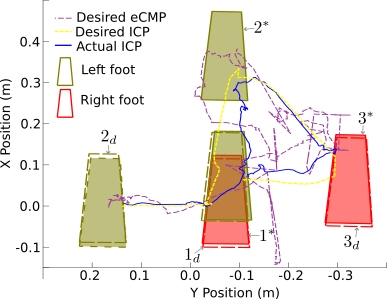}
    \vspace{-2mm}
    \caption{Data from the push shown in Fig. \ref{fig:nadia_crossover_recovery}. Robot was walking in place, with the achieved footholds as filled, solid lines and shown as $[\cdot]^*$, and the desired footholds as dashed.}
    \label{fig:nadia_crossover_data}
\end{figure}

\begin{figure*}[t!]
    \centering
    \includegraphics[width=0.85\textwidth]{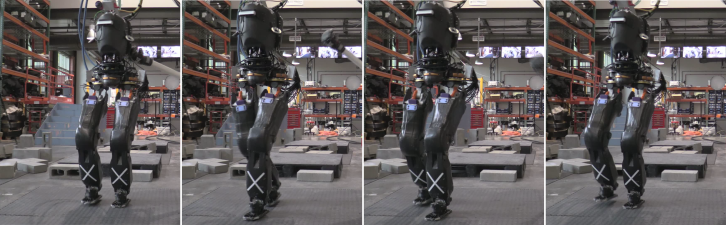}
    \vspace{-2mm}
    \caption{Nadia recovering from a push to the inside of a step. The robot crosses its legs using the reachability constraints in Fig. \ref{fig:crossover_reachability}, and continues walking.}
    \vspace{-6mm}
    \label{fig:nadia_crossover_recovery}
\end{figure*}

To test the benefits of using cross-over steps on the robot hardware, we directed the robot to walk in place and applied disturbances to the torso towards the inside of the step. 
Images from this experiment can be seen in Fig. \ref{fig:nadia_crossover_recovery}, with data in Fig. \ref{fig:nadia_crossover_data}.
After the disturbance, the robot crossed the swing leg in front of the stance leg within the reachability constraint, and then adjusted outward on the subsequent step, to then resume stepping in place.
Additionally, in Fig. \ref{fig:nadia_crossover_data}, it can be seen that the feedback eCMP saturates  to a maximum distance outside the right side of the support polygon to help maintain stability.
If the robot had been unable to perform the cross-over step, the left recovery step would have been placed several (approx. 10) centimeters further left, requiring an extremely large subsequent step, as every additional centimeter the foot moves requires $10cm$ of step adjustment based on the $~0.73s$ swing duration used.
From this, it is highly unlikely that the robot would have been able to recover without the use of cross-over steps.

\section{Conclusion}
Step adjustment is a critical component for dynamic locomotion of humanoid robots. 
In this work, we present a novel approach for enabling step adjustment on humanoid hardware, including cross-over steps, as well as a better understanding of how reachability constraints affect the ability to regain balance.
We also propose a novel method for varying the transfer duration for stability, as well as using the hip and ankle strategy for balance.
From this approach, it becomes apparent that constraining the footstep adjustment significantly affects the recovery dynamics, and should be accounted for.
We also show from hardware tests and software simulations that the inclusion of cross-over steps enables a much greater ability to recover from inward disturbances.
Transfer adjustment is also extremely beneficial for recovering from large disturbances. 
In the future, we hope to include environmental constraints, such that the robot can perform step adjustment when navigating complex terrain.

\bibliography{mybib}

\end{document}